\PassOptionsToPackage{dvipsnames}{xcolor}

\documentclass{article} 
\usepackage{iclr2027_conference,times}

\usepackage{amsmath,amsfonts,bm}

\def\eqref#1{equation~\ref{#1}}

\def\1{\bm{1}}

\DeclareMathAlphabet{\mathsfit}{\encodingdefault}{\sfdefault}{m}{sl}
\SetMathAlphabet{\mathsfit}{bold}{\encodingdefault}{\sfdefault}{bx}{n}

\usepackage{hyperref}
\usepackage{url}
\usepackage{algorithm,algpseudocode}
\usepackage{xspace}
\usepackage{amssymb} 
\usepackage{booktabs}
\usepackage{graphicx}
\usepackage{multirow}
\usepackage{pifont}
\usepackage{xcolor}

\newcommand{\xmark}{\ding{55}}%

\title{Beyond Corrected Actions: Training LLM Agents on Textual Critiques}

\usepackage[dvipsnames]{xcolor}

\usepackage[most]{tcolorbox}
\usepackage{fancyvrb}
\usepackage[varqu,varl,scaled=1.0]{inconsolata}

\definecolor{promptrule}{HTML}{31506E}
\definecolor{promptbg}{HTML}{F4F7FB}
\definecolor{interjrule}{HTML}{8A5A2B}
\definecolor{interjbg}{HTML}{FDF6EE}
\definecolor{minrule}{HTML}{3F6B4A}
\definecolor{minbg}{HTML}{F3F9F4}

\tcbset{
  promptstyle/.style={
    enhanced, breakable=false,
    coltitle=white,
    fonttitle=\bfseries\footnotesize,
    boxrule=0.6pt, arc=2pt,
    left=5pt, right=5pt, top=3pt, bottom=3pt,
    before skip=0pt, after skip=4pt,
    bottomtitle=1.5pt, toptitle=1.5pt,
  }
}

\newtcolorbox{promptbox}[1][]{promptstyle,
  colback=promptbg, colframe=promptrule, title={#1}}
\newtcolorbox{minpromptbox}[1][]{promptstyle,
  colback=minbg, colframe=minrule, title={#1}}
\newtcolorbox{interjbox}[1][]{promptstyle,
  colback=interjbg, colframe=interjrule, title={#1}}

\fvset{listparameters={\setlength{\topsep}{0pt}\setlength{\partopsep}{0pt}}}
\DefineVerbatimEnvironment{promptverb}{Verbatim}
  {fontsize=\scriptsize, baselinestretch=0.88, xleftmargin=0pt,
   xrightmargin=0pt, samepage=true}

\author{
Thomas Zeng\textsuperscript{1$\dagger$},
Zelei Cheng\textsuperscript{2},
Mingyang Zhou\textsuperscript{2},
Sambit Sahu\textsuperscript{2},
William Campbell\textsuperscript{2},\\
\textbf{
Genta Indra Winata\textsuperscript{2}}
\\[0.5em]
{\normalfont\small
\textsuperscript{1}University of Wisconsin--Madison
\textsuperscript{2}AI Foundations, Capital One
}
}

\begingroup

\footnotetext[2]{Work performed during an internship at Capital One. Correspondence to Thomas Zeng at \texttt{tpzeng@wisc.edu} and Genta Winata at \texttt{genta.winata@capitalone.com}.}
\endgroup

\newcommand{\method}{CAFT\xspace} 
\xspaceaddexceptions{+}

\algtext*{EndFor}
\algtext*{EndIf}
\iclrfinalcopy

\begin{document}

\maketitle

\begin{abstract}
Large language models (LLMs) are increasingly used for multi-turn agentic tasks that require interactions with users and environments over extended sequences of actions. A common approach to improving LLM agents is to use more capable expert models to provide training supervision, either by behavior cloning the expert's trajectories or by using the expert to correct the actions in trajectories generated by the weaker student model.
However, expert LLMs can provide richer supervision than actions alone: they can directly generate textual critique that explain \emph{why} specific actions are incorrect. We study whether training on such feedback benefits LLM agents and propose \textbf{\underline{C}}ritique-Augmented \textbf{\underline{A}}gentic \textbf{\underline{F}}ine-\textbf{\underline{T}}uning (\method), a framework that elicits textual expert feedback on student-generated trajectories. For each incorrect action, the expert yields both a corrected action \emph{and} a textual critique explaining the student's mistake---which we then use as the training target for fine-tuning. Across four student models from the Nemotron 3 and Gemma 4 families on the agentic domains $\tau^2$-Bench Airline+Retail and WebShop, adding critiques to action-level supervision improves performance by 8\% on average over action-only distillation. Additionally, \method checkpoints can also provide a better warm start for reinforcement learning than expert-trajectory distillation. Our results demonstrate that textual critiques can strengthen supervision for multi-turn LLM agents.
\end{abstract}

\section{Introduction}
\label{intro}

\begin{figure}[t]
\centering
\includegraphics[width=\textwidth]{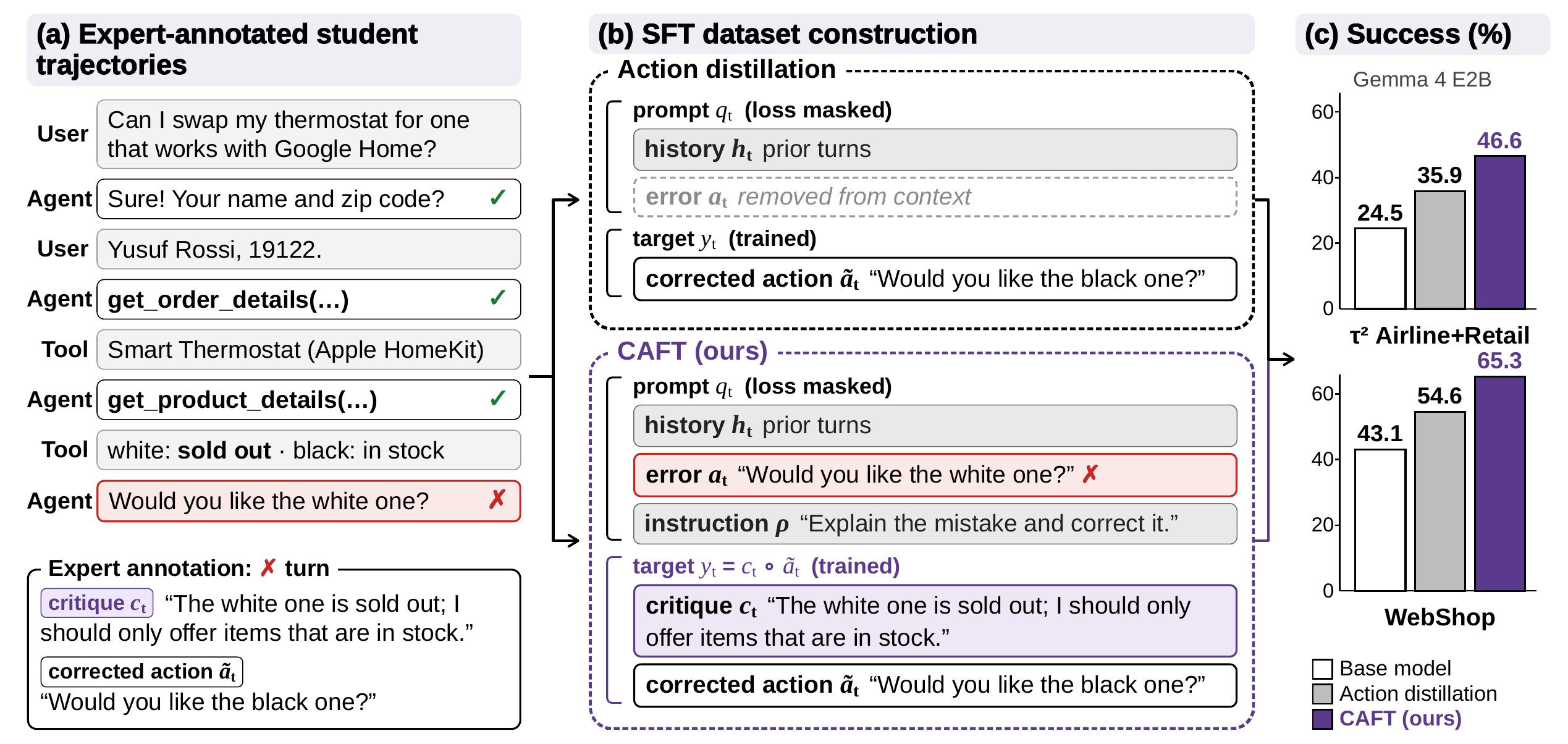}
\caption{\textbf{Overview of \method.}
(a) The student model is rolled out and an expert labels every agent turn, writing a critique $c_t$ and a corrected action $\tilde a_t$ for each incorrect turn.
(b) Action distillation replaces the incorrect step with $\tilde a_t$ while \method keeps the incorrect step in context and trains on the critique followed by the corrected action.
(c) Task success rate of Gemma 4 E2B on $\tau^2$ Airline+Retail and WebShop.}
\label{fig:overview}
\end{figure}

Large language models (LLMs) are increasingly deployed as agents in multiturn tool-using settings that require interacting with users and environments over extended sequences of decisions \citep{mohammadi2025evaluation,pan2025measuring,xu2026theagentcompany,winata2026t1}. 
Success in these settings requires correctly executing a sequence of interdependent actions.
Importantly, agent actions can be incorrect for many  different reasons; such as violating a domain policy, issuing an invalid tool call or acting before having collected all the information it needs.  Therefore, training reliable multi-turn agents requires supervision that steers them away from this diverse set of failures.

When a stronger expert LLM is available, the simplest way to improve a smaller student agent is behavior cloning the expert's successful trajectories \citep{chen2024agent,zeng2024agenttuning,chen2025atlas}. However, expert trajectories primarily demonstrate how the expert succeeds and provide little direct supervision on the mistakes that a weaker student itself tends to make.  DAgger-style \citep{ross2011reduction} methods thus instead obtain expert supervision on states the student visits. Recent work has adapted this idea directly to long-horizon LLM agents by relabeling learner-relevant states with expert actions \citep{li2026revisiting}, or by providing dense token-level teacher supervision on sequences generated by the student itself \citep{agarwal2024policy, liao2026multi}. 

While these approaches supervise what action the expert would execute, they do not provide supervision on why the student’s original action is incorrect. 
Yet as the expert is itself an LLM, 
it \emph{can} directly make this explanation explicit by identifying the specific policy, information requirement, or tool-use constraint violated by the student's action.
This capability makes it possible to construct training methods that incorporate the expert's textual critique of the student's mistake in addition to the corrected action. 
As this textual critique can convey information not captured by the corrected action alone, we hypothesize that it provides additional learning signal, and ask:
\begin{quote}
     \textit{Does training an LLM agent on the expert's critique of its incorrect actions improve the policy beyond training on the corrected expert action alone?}
\end{quote}

To study this question, we introduce Critique-Augmented Agentic Fine-Tuning (\method). \method first rolls out the student policy to collect the states and mistakes it encounters under its own trajectory distribution. A privileged expert model then reviews these trajectories, identifies incorrect student actions, and provides both a corrected action and a natural-language critique explaining what was wrong with the student's original action. 
\method then creates a fine-tuning dataset that keeps each erroneous action in context and trains the student to produce both the critique and the counterfactually corrected action generated by the expert.

We evaluate \method on two benchmarks: WebShop \citep{yao2022webshop} and the Airline and Retail domains of $\tau^2$-Bench \citep{barres2026taubench}. Across
four student models from the Nemotron 3 and Gemma 4
families \citep{blakeman2025nvidia, team2026gemma},
\method consistently improves over action-only supervision on student trajectories by $8$\% on average and by up to $27$\% on WebShop with Nemotron 3 Nano 4B. We further find in a single-model case study that \method checkpoints provide a stronger initialization for reinforcement learning (RL) than either the base model or standard expert-model distillation. Our contributions are summarized as follows:
\begin{itemize}
 \item We introduce Critique-Augmented Agentic Fine-Tuning (\method), which trains LLM agents on their own trajectories annotated with expert critiques explaining \emph{why} a mistake occurred in addition to the corrected actions.
 \item We show that \method consistently improves over training on corrected actions alone across four student models and two agentic environments. Furthermore, targeted ablations show that the textual critiques provide a stronger learning signal than the corrected actions alone.
 \item We demonstrate the importance of privileged information in generating high quality critique generation, compare \method with expert-trajectory distillation and repeated fine-tuning, and evaluate it as an initialization for RL.
\end{itemize}

\section{Training LLM Agents on Textual Critiques}
\label{methods}

\begin{algorithm}[t]
\caption{\method: Constructing critique-augmented supervision from student trajectories }
\label{alg:critique-targets}
\begin{algorithmic}[1]
\Require Task set $\mathcal{X}$, base policy $\pi_0$, environment $\mathcal{E}$, expert $C$, interjection $\rho$
\State $\mathcal{D}_{\mathrm{SFT}} \gets \varnothing$
\For{$x \in \mathcal{X}$}
    \State $\mathcal{T}_0(x) \gets \Call{CollectTrajectories}{\pi_0,\mathcal{E},x}$
    \For{$\tau \in \mathcal{T}_0(x)$}
        \For{$t=1,\ldots,T(\tau)$}
            \State $(z_t,c_t,\tilde a_t) \gets C(h_t,a_t;x)$
            \If{$z_t = 1$}
                \State $(q_t,y_t) \gets (h_t,a_t)$
            \Else
                \State $q_t \gets h_t \circ a_t \circ \rho$
                \State $y_t \gets c_t \circ \tilde a_t$
            \EndIf
            \State $\mathcal{D}_{\mathrm{SFT}}
                \gets
                \mathcal{D}_{\mathrm{SFT}} \cup \{(q_t,y_t)\}$
        \EndFor
    \EndFor
\EndFor
\State \Return $\mathcal{D}_{\mathrm{SFT}}$
\end{algorithmic}
\end{algorithm}
To provide agents with explicit supervision on why their actions are incorrect, \method fine-tunes a student policy $\pi_0$ using expert critiques constructed from the student's own trajectories (refer to Figure~\ref{fig:overview} for an overview). We first roll out with $\pi_0$ on a collection of multi-turn agentic tasks, and then use an expert $C$ to evaluate each agent turn. For turns judged incorrect, the expert provides both a natural-language critique and the corrected action it would have taken instead. These annotations are converted into supervised examples used to fine-tune a policy $\pi_\theta$ from $\pi_0$. In our experiments, $C$ is an LLM with access to privileged task information, such as reference actions and evaluation criteria, which grounds its judgments in task
requirements unavailable to the student.
The complete procedure for constructing this training dataset is outlined in Algorithm~\ref{alg:critique-targets}.

\paragraph{Multi-turn agent setting.}
To formally represent the multi-turn agentic setting that our method focuses on, we define the agent interaction as a sequential decision making process.
Let $\mathcal{X}$ denote a collection of task instances and let $x\sim p_{\mathcal X}$ denote a sampled task. Given $x$, the environment $\mathcal E$ emits an initial agent-visible observation $o_1$ describing the task. At agent turn $t$, the policy generates an action based on the previous interaction history
\[
    a_t \sim \pi(\cdot \mid h_t),
    \qquad \text{where }
    h_t=(o_1,a_1,\ldots,o_t),
\]
after which the environment $\mathcal E$ produces the next observation $o_{t+1}$. Here, an action $a_t$ is the agent output at turn $t$ (e.g., a tool call or a user-facing message), while $o_t$ is the information returned to the agent (e.g., a tool call result or a user reply). A rollout therefore produces a trajectory
\[
    \tau=(o_1,a_1,o_2,a_2,\ldots,o_T,a_T, o_{T+1}).
\]
We denote by
$p_{\pi,\mathcal E}(\tau\mid x)$
the trajectory distribution induced by policy $\pi$ and environment $\mathcal E$ on task $x$.
Each task is evaluated by a binary correctness grader
$R(x,\tau)\in\{0,1\},$
where $R(x,\tau)=1$ indicates successful completion. The agent's objective is to maximize expected task success,
\[
    J(\pi)
    =
    \mathbb{E}_{x\sim p_{\mathcal X}}
    \mathbb{E}_{\tau\sim p_{\pi,\mathcal E}(\cdot\mid x)}
    \left[R(x,\tau)\right].
\]
\method aims to improve $J$ by constructing supervised training data from trajectories sampled from the base policy $\pi_0$.

\paragraph{Expert annotations.}
We now precisely describe the expert annotation process. For each trajectory $\tau\sim p_{\pi_0,\mathcal E}(\cdot\mid x)$, the expert model $C$ evaluates every agent turn $a_t$. It first assigns a correctness label
\[
    z_t\in\{0,1\},
\]
where $z_t=1$ denotes a correct action and $z_t=0$ an incorrect action. When $z_t=0$, the expert additionally generates a natural-language critique $c_t$ explaining the error and a corrected action $\tilde a_t$ specifying what the agent should have done given $h_t$. We summarize this annotation as
\[
    (z_t,c_t,\tilde a_t)=C(h_t,a_t;x),
\]
where $c_t$ and $\tilde a_t$ are required only when $z_t=0$. In our implementation, $C$ may additionally receive privileged information associated with task $x$ that is unavailable to the agent, such as gold reference actions or the evaluation criteria. Both the critique and correction are generated in the agent's output format so that they can be used directly as supervised targets. For more details on the specific expert prompts used, refer to Appendix~\ref{app:prompts}.

\paragraph{Supervised fine-tuning dataset construction.}
For the construction of the fine-tuning dataset, we convert each annotated agent turn into an independent supervised training example $(q_t,y_t)$ where $q_t$ is the context containing the preceding agentic interactions and $y_t$ is the response the student is trained to produce.
Let $\rho$ denote a fixed natural-language interjection inserted as a user message, indicating that the preceding action was incorrect and requesting an explanation and a corrected next action. We use $\circ$ to denote concatenation under the model's chat template.
For a correct turn ($z_t=1$), we train the model to reproduce the action originally sampled by $\pi_0$.
Thus, correct behavior from the base policy is retained as ordinary behavior-cloning supervision---which we include because our ablation experiments (see Appendix~\ref{app:passweight}) show they are beneficial. For an incorrect turn ($z_t=0$), we retain the erroneous action $a_t$ in the context, append the interjection $\rho$, and train the model to produce the expert's critique followed by its corrected action.
The training example associated with turn $t$ is thus
\[
(q_t,y_t)
=
\begin{cases}
    (h_t,\;a_t),
        & z_t=1,\\[4pt]
    (h_t\circ a_t\circ\rho,\;
     c_t\circ\tilde a_t),
        & z_t=0.
\end{cases}
\]
The resulting supervised dataset is
\[
    \mathcal D_{\mathrm{SFT}}
    =
    \bigcup_{x\in\mathcal X}
    \bigcup_{\tau\in\mathcal T_0(x)}
    \bigcup_{t=1}^{T(\tau)}
    \left\{(q_t,y_t)\right\},
\]
where $\mathcal T_0(x)$ denotes the set of trajectories collected from $\pi_0$ on task $x$.
Finally, we fine-tune $\pi_\theta$ on $\mathcal D_{\mathrm{SFT}}$ using the standard autoregressive language-modeling objective. Writing $y=(y_1,\ldots,y_{|y|})$ for the target tokens, we minimize
\[
    \mathcal L_{\mathrm{SFT}}(\theta)
    =
    -\sum_{(q,y)\in\mathcal D_{\mathrm{SFT}}}
    w(q,y)
    \sum_{j=1}^{|y|}
    \log
    \pi_\theta
    \left(
        y_j
        \mid
        q\circ y_{<j}
    \right).
\]
The weight $w(q,y)$ is $1$ on incorrect turns and $\lambda$ on correct turns (where we use $\lambda=0.1$ throughout; chosen following the results of a sweep in Appendix~\ref{app:passweight}).
Lastly, we note the expert, $\rho$, and the critiques appear only in the training data. At inference time, the finetuned policy acts as an ordinary agent, and is \emph{not} interjected by critique from an expert model or asked to critique itself.

\section{Experiments}
\label{experiments}

\begin{table}[t]
\centering
\small
\setlength{\tabcolsep}{3.8pt}
\caption{
\textbf{Comparison of \method and its ablated variants against action distillation.}
Task success (\%) on $\tau^2$ AR and WebShop for the base policies, the action-distillation baseline, and \method and its ablations. \method-Critique and \method consistently outperform all other methods in all student--environment pairs.
}
\label{tab:critique_main}

\begin{tabular}{llrrrrr}
\toprule
Environment
& Student Model
& Base
& \multicolumn{1}{c}{Action}
& \multicolumn{3}{c}{\method} \\
\cmidrule(lr){5-7}
&
&
&
Distillation
& Correction
& Critique
& Full \\
\midrule

\multirow{3}{*}{$\tau^2$ AR}
& Gemma 4 E2B
    & 24.5
    & 35.9
    & 37.0
    & 42.6
    & \textbf{46.6} \\

& Gemma 4 E4B
    & 42.4
    & 49.7
    & 49.3
    & \textbf{53.2}
    & 52.1 \\

& Gemma 4 31B
    & 65.1
    & 66.7
    & 67.0
    & 69.4
    & \textbf{69.4} \\
& Nemotron 3 Nano 4B
    & 47.8
    & 52.1
    & 57.6
    & 59.3
    & \textbf{60.3} \\

\midrule

\multirow{3}{*}{WebShop}
& Gemma 4 E2B
    & 43.1
    & 54.6
    & 55.9
    & 60.3
    & \textbf{65.3} \\

& Gemma 4 E4B
    & 42.1
    & 59.2
    & 58.0
    & 59.5
    & \textbf{62.6} \\

& Gemma 4 31B
    & 53.1
    & 59.6
    & 57.9
    & 60.4
    & \textbf{61.0} \\
& Nemotron 3 Nano 4B
    & 24.3
    & 31.6
    & 37.1
    & 54.8
    & \textbf{58.9} \\

\bottomrule
\end{tabular}
\end{table}

In this section, we introduce the experiments and ablation studies that we conducted to validate the design of \method. The details of the experiment setup is described in Secion~\ref{sec:exp_setup}
In Section~\ref{sec:critique_main}, we establish the main claim that textual critique carries supervision beyond the corrected action alone when training LLM agents.
Section~\ref{sec:critique_teacher} then asks what the expert needs in order to produce useful critiques, and shows that privileged task information matters more than fine-grained prompting.
Sections~\ref{sec:hybrid_distillation} and places
\method against the natural alternatives---distilling the expert's trajectories, and \ref{sec:round2} studies the effectiveness of running mulitple iterations of \method. Finally, Section~\ref{sec:rl_warmstart} asks whether \method can be used effectively as a warm start for RL.
\subsection{Experimental Setup}
\label{sec:exp_setup}

\paragraph{Environments and evaluation.}
We evaluate on the Airline and Retail domains of $\tau^2$-Bench ($\tau^2$ AR) \citep{barres2026taubench}, and WebShop \citep{yao2022webshop}.
$\tau^2$~AR evaluates conversational agents that must interact with a user and external tools while following domain-specific policies, on tasks such as booking a flight or modifying an online order.
WebShop evaluates agents that navigate a simulated online store to find and purchase
an item satisfying a natural-language request.
For $\tau^2$ AR, we use the benchmark's outcome-based evaluator, and for WebShop, we report the fraction of episodes that purchase an item satisfying all the request's constraints. For both datasets, we collect $16$ rollouts per task.
Full details are in Appendix~\ref{app:envs}.

\paragraph{Models and data collection.}
We evaluate four student policies: NVIDIA Nemotron 3 Nano 4B, Gemma 4 E2B, Gemma 4 E4B and Gemma 4 31B \citep{blakeman2025nvidia,team2026gemma}.
We use Gemma 4 31B as the expert model for generating turn-level correctness judgments, critiques, and corrected actions.
When Gemma 4 31B is also the
student, the expert critiques its own trajectories, so this setting tests whether a model can improve
from \emph{self}-critique given privileged task information.
We partition both environments into training and evaluation task splits. Then, for each student--environment pair we roll out the base policy $\pi_0$ over the training tasks with $16$ rollouts per task and have the expert annotate every agent turn.
Complete collection, dataset-construction, and optimization settings are given in
Appendix~\ref{app:sft}.

\paragraph{Baselines and supervision variants.}
We compare \method against the unmodified base policy, two ablations of \method and an action distillation baseline designed to isolate the supervision typified by LEAP \citep{choudhury2025better}, SOPD \citep{sun2026step}, and \citet{ye2026few}.
Our experiment controls for both the student-generated trajectories and the expert annotations, varying \emph{only} the supervision used for expert-flagged turns:
\begin{itemize}
    \item \textbf{Base}: the original student policy $\pi_0$, without
    additional fine-tuning.

    \item \textbf{Action Distillation}: trains the student on the
    expert's corrected action alone, with the mistake removed from context,
    $(q_t,y_t)=(h_t,\tilde a_t)$.

    \item \textbf{\method-Correction}: removes the critique target from
    \method, keeping only the expert correction,
    $(q_t,y_t)=(h_t\circ a_t\circ\rho,\tilde a_t)$.

    \item \textbf{\method-Critique}: removes the corrected-action
    target from \method and trains only on the expert's textual critique,
    $(q_t,y_t)=(h_t\circ a_t\circ\rho,c_t)$.

    \item \textbf{\method}: trains on both the expert's critique and corrected
    action,
    $(q_t,y_t)=(h_t\circ a_t\circ\rho,\,
    c_t\circ\tilde a_t)$.
\end{itemize}

\subsection{Do Critiques Provide Supervision Beyond Corrected Actions?}
\label{sec:critique_main}

We compare \method against the baselines and the two variants of \method and summarize the results in Table~\ref{tab:critique_main}. 
First, we observe that training on correct expert actions yields clear improvements over the base model.
The Action Distillation baseline consistently outperforms the base model across all settings. However, incorporating textual critique via \method gives further improvements. \method outperforms Action Distillation in all eight settings, by 8.4\% on average, and delivers the strongest overall performance in six of the eight cases.

To isolate the value of the textual critique, we compare our two \method ablations. Crucially, \method-Critique never trains on the corrected action for flagged turns, yet it consistently outperforms \method-Correction. This provides direct evidence that explaining the error is more beneficial to the student model than merely providing the correct action. We further observe that \method-Correction, which differs from Action Distillation primarily in keeping the mistaken step in the context, performs similarly to the Action Distillation baseline in most settings. Thus, the majority of \method's advantage stems directly from the textual critique.
Comparing the two \method ablations, \method-Critique consistently outperforms \method-Correction across all student–environment pairs.
While gains vary considerably by setting, the average is 5.0\%, with the largest from Nemotron 3 Nano 4B on WebShop.

Finally, we highlight the results for the Gemma 4 31B experiments where the expert is the student itself. In this setting, \method still improves over the base model (by $4.3$\% on $\tau^2$ AR and $7.9$\% on WebShop) and over Action Distillation, albeit by smaller margins. This suggests that \method may remain useful when no stronger expert is available. 
In summary, the results demonstrate that the textual critiques provide effective supervision to improve LLM agents.
Training student agents to predict critiques over their own mistakes
gives non-trivial gain over using only the correct actions, establishing \method as a broadly effective supervision paradigm for fine-tuning LLM agents.

\subsection{How Much Does the Expert Need to Know to Critique Effectively?}
\label{sec:critique_teacher}

The expert prompt used for the experiments in Section~\ref{sec:critique_main} gives the expert two types of help.
The first is a detailed domain-specific prompt describing what types of actions are incorrect; i.e., actions that apply to the wrong entity, skip a policy-mandated verification, perform an incorrect or irreversible write, or contradict the policy or the user.
The second is privileged task information such as the gold reference actions, the user scenario and the evaluation criteria used to that the student never sees (Appendix~\ref{app:prompts}).

To isolate the contribution of each, 
we construct an ablation on the three smaller student models that independently varies whether the
expert receives a detailed instruction or privileged task
information. We re-annotate the student trajectories with the same expert under each condition and rerun \method.
Table~\ref{tab:teacher_ablation} shows that privileged information is the dominant factor. Removing it reduces
success by 7.5\% on average.
In contrast, when privileged information is available, replacing the detailed
instruction with a minimal one retains most of the performance.
The detailed prompt matters most for the weakest model, Gemma 4 E2B, while having no impact on Nemotron 3 Nano 4B. In summary, high-quality supervision from the expert model therefore primarily depends on the expert knowledge on the task requirements instead of  the detailed guidance.

\begin{table}[t]
\centering
\small
\caption{
\textbf{Ablating privileged information and detailed guidance in the expert prompt.}
$\tau^2$ AR
task success (\%) of CAFT trained on critiques from Gemma 4 31B, varying whether the expert receives privileged task information and a detailed critique prompt. 
Privileged information has a larger impact than the detailed prompt.
}
\label{tab:teacher_ablation}
\begin{tabular}{cccc c}
\toprule
Detailed & Privileged & Gemma 4 & Gemma 4 & Nemotron 3 \\
prompt   & info       & E2B     & E4B     & Nano 4B    \\
\cmidrule(lr){1-2} \cmidrule(lr){3-5}
\checkmark & \checkmark & \textbf{46.6} & \textbf{52.1} & 60.3 \\
\checkmark & \xmark     & 34.2          & 44.7          & 57.7 \\
\addlinespace
\xmark     & \checkmark & 40.1          & 50.0          & \textbf{60.7} \\
\xmark     & \xmark     & 32.9          & 43.3          & 57.2 \\
\bottomrule
\end{tabular}
\end{table}

\subsection{Comparison to Expert-Trajectory Distillation}
\label{sec:hybrid_distillation}

We also compare \method to direct distillation from expert trajectories on agentic tasks to validate the idea of  training students LLM agent to correct their own mistakes is a stronger alternative than just imitating teacher's behaviors.  
Our expert model, Gemma 4 31B, outperforms every base student, achieving
65.1\% and 53.1\% task success on $\tau^2$ AR and WebShop, respectively (Table~\ref{tab:critique_main}).
So its trajectories are a natural source of behavior-cloning data.
Table~\ref{tab:expert_distill} compares this Expert Distillation
baseline with \method for the three smaller students.

On WebShop, expert distillation proves entirely ineffective, bringing success rate to around 15.0--19.1\% (below the base model's 24.3-43.1\%) despite the expert outperforming every student model.
This is in contrast to the strong performance of \method.
Inspecting the distilled policy trajectories, we observe they continuously search for new items until they exhaust the allowed turn budget, without committing to a purchase. This suggests the models learned to mimics the expert's exploration patterns but not its stopping conditions. \method avoids this collapse because its supervision is
anchored on the student's own context and mistakes.

On $\tau^2$ AR, the results are more nuanced. Expert distillation is slightly better than
\method on Gemma 4 E4B and Gemma 4 E2B, while \method performs better on Nemotron 3 Nano 4B.
Given expert trajectories help on $\tau^2$ AR but
cause degenerate behavior on WebShop, we investigate whether the two
supervision sources can be complementary for $\tau^2$ AR. 
The \method+BC variant mixes 50\% expert trajectories for behavior cloning (BC) with 50\% \method data. 
This mixture beats both of its components for
both Gemma 4 E2B (53.7\%) and
Gemma 4 E4B (60.1\%). For Nemotron 3 Nano 4B, however, \method+BC performs
worse than \method alone (56.0\% vs. 60.3\%).
Taken together, expert demonstrations can therefore complement critique supervision, but the benefit depends on the student and environment. The expert’s task success alone is
not sufficient to  determine whether its demonstrations will improve the student.

\begin{table}[t]
\centering
\small
\caption{
\textbf{Comparison of \method with expert-trajectory distillation.}
Task success (\%). Expert
Distillation behavior-clones trajectories from Gemma 4 31B; \method+BC mixes 50\% expert trajectories with 50\% CAFT data. Expert distillation collapses on WebShop, so \method+BC is not run
there.}
\label{tab:expert_distill}

\begin{tabular}{llrrr}
\toprule
Environment & Student & Expert Distill. & \method & \method+BC \\
\midrule

\multirow{3}{*}{$\tau^2$ AR}
& Gemma 4 E2B
    & 50.4
    & 46.6
    & \textbf{53.7} \\
& Gemma 4 E4B
    & 53.8
    & 52.1
    & \textbf{60.1} \\
& Nemotron 3 Nano 4B
    & 53.3
    & \textbf{60.3}
    & 56.0 \\

\midrule

\multirow{3}{*}{WebShop}
& Gemma 4 E2B
    & 19.1
    & \textbf{65.3}
    & --- \\
& Gemma 4 E4B
    & 16.4
    & \textbf{62.6}
    & --- \\
& Nemotron 3 Nano 4B
    & 15.0
    & \textbf{58.9}
    & --- \\

\bottomrule
\end{tabular}
\end{table}

\subsection{Does a Second Iteration of \method Help?}
\label{sec:round2}

To investigate the upper bounds of our approach, we explore whether the benefits of \method compound across multiple iterations or if the performance quickly saturates. 
Specifically, we experiment with taking the checkpoint from the first iteration and then rolling out and re-annotating new trajectories to fine-tune for a second iteration.

Table~\ref{tab:round2} shows the result of a second iteration of training. From it, we see a clear difference by environment. On WebShop, a second iteration improves every
student by $5.3$\% on average and by as much as $10.9$\% for Nemotron 3 Nano 4B.
On $\tau^2$ AR the
picture is mixed: while the two small Gemma students improve modestly ($1.6$\% and $3.1$\%), Nemotron 3
Nano 4B does not see any more improvements, suggesting that the gains saturated after one iteration.
We thus find that in general another iteration of \method \emph{can} help. Although the amount it helps depends strongly on the environment and headroom after the first iteration.

\begin{table}[t]
\centering
\small
\caption{
\textbf{Using \method for a second iteration.}
Task success (\%) after one and two iterations of \method. A second iteration helps every student on
WebShop and the smaller Gemma students on $\tau^2$ AR.
}
\label{tab:round2}
\begin{tabular}{llrrr}
\toprule
Environment & Student Model & iteration 1 & iteration 2 \\
\midrule

\multirow{4}{*}{$\tau^2$ AR}
& Gemma 4 E2B        & 46.6 & \textbf{48.2}  \\
& Gemma 4 E4B        & 52.1 & \textbf{55.2}  \\
& Gemma 4 31B        & \textbf{69.4} & 65.4  \\
& Nemotron 3 Nano 4B & \textbf{60.3} & 58.8  \\
 \midrule

\multirow{4}{*}{WebShop}
& Gemma 4 E2B        & 65.3 & \textbf{72.1} \\
& Gemma 4 E4B        & 62.6 & \textbf{64.9} \\
& Gemma 4 31B        & 61.0 & \textbf{62.1}  \\
& Nemotron 3 Nano 4B & 58.9 & \textbf{69.8}  \\
 \bottomrule
\end{tabular}
\end{table}
\subsection{Can We Use \method as a Warm Start for RL?}
\label{sec:rl_warmstart}

Supervised fine-tuning is commonly used to warm start reinforcement learning (RL), placing the policy in a higher-reward region to improve sample efficiency \citep{ouyang2022training, guo2025deepseek}.
Canonically, warm-starting is achieved by distilling trajectories from a stronger model. We thus ask if \method yields a better initialization than expert distillation.

As an initial investigation, we use Nemotron 3 Nano 4B on $\tau^2$ AR as a case study. We run 300 steps of GRPO \citep{shao2024deepseekmath} on the $
\tau^2$ AR training split, varying only the initialization: (i) the base model (ii) the expert distillation checkpoint and (iii) the \method checkpoint. All three runs share the same GRPO hyperparameters, with full details in Appendix \ref{app:rl}. As shown in Figure~\ref{fig:rl}, \method outperforms both alternatives by around 5\% on the held-out tasks.
Notably, while the expert distillation warm-start achieves a higher training reward than the base model, it does not transfer---with its validation performance collapsing to the base model.

Since both warm starts are trained on the same training tasks that RL later optimizes,
we hypothesize that 
behavior cloning on the expert’s solutions to these tasks instills task specific patterns, which RL then amplifies. This then raises training reward without improving generalization.
Meanwhile, \method's supervision instead targets the student's own mistakes, which appears to transfer better to unseen tasks. Lastly, we note \method also makes RL more sample-efficient: the \method initialized run matches the base-initialized run’s best held-out success after about 60 of 300 steps.

\begin{figure}[t]
\centering
\includegraphics[width=\textwidth]{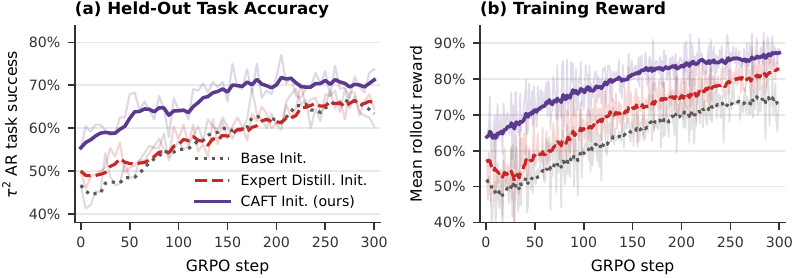}
\caption{\textbf{Using \method as a warm start for RL.}
GRPO on $\tau^2$ AR with Nemotron 3 Nano 4B from three initializations.
(a) Validation task success, averaged over the airline and retail validation splits. (b) Mean reward of on-policy rollouts during training.}
\label{fig:rl}
\end{figure}

\section{Related Work}

\paragraph{Expert supervision for training LLM agents.}
A common approach to train LLM agents is to imitate the behavior from the trajectories generated by stronger models~\citep{zeng2024agenttuning, chen2024agent, chen2025atlas}. However at inference time, a weaker student may encounter states absent from the expert data. DAgger addresses this problem by obtaining expert labels on states induced by the student policy~\citep{ross2011reduction}. Analogously, on-policy distillation trains on student-generated sequences with dense token-level supervision from a teacher model~\citep{agarwal2024policy}. Recent work has extended these ideas to long-horizon LLM agents~\citep{li2026revisiting, liao2026multi} and to \emph{action-level} supervision on student-visited states, where a privileged teacher rewrites the student's erroneous step \citep{choudhury2025better, lyustudent}, or a teacher continues from a student prefix for one \citep{sun2026step} or a few \citep{ye2026few} steps. However, all the existing approaches directly drop the student's mistakes from the training context by replacing them with the teacher's action. Unlike the previous methods, 
\method shares the goal of supervising the states
the student itself visits (our Action Distillation baseline, Section~\ref{sec:exp_setup}, instantiates this form of supervision), but keeps the mistake in context so that we can train on the expert’s critique of it.

\paragraph{Training from natural language critiques.}
Several works have explored using critiques as supervision for language models. In single-turn reasoning, Critique Fine-Tuning (CFT) trains models to critique noisy responses rather than imitate correct solutions \citep{wang2025critique}.
LEMA fine-tunes on an explanation of an incorrect solution followed by its correction \citep{an2023learning} and \cite{ye2025physics} show how error-correction data can be used for mathematical pretraining.
Critique-Guided Distillation (CGD) instead trains the student to refine its own response \emph{given} a teacher's critique \citep{kapusuzoglu2026critiqueguided}. Related reinforcement-learning approaches likewise use natural-language critiques to guide policy improvement \citep{zhang2026advancing,lin2026icrl}.
For agents, Early Experience  trains a policy to write its own reflection on why an expert action
beats an alternative at expert-visited states \citep{zhang2026agent}, and CAST conditions a tool-calling policy on a separate critic’s verdicts \citep{saeidi2026cast}.
\method builds on these works by asking whether training a multi-turn agent to explain its own
mistakes improves its decisions beyond learning the expert's corrected
actions alone.

\paragraph{Reflection and error recovery in LLM agents.}
A complementary line of work uses natural language feedback to enable LLM agents to reflect and self-correct. Reflexion uses verbal feedback stored in an episodic memory to improve subsequent attempts without updating model parameters \citep{shinn2023reflexion}. Critique-Guided Improvement (CGI) trains a separate critic to generate actionable natural-language feedback and an actor to revise its candidate action conditioned on the critic's generated feedback \citep{yang2025lighthouse}.
Other work directly trains agents to exhibit reflection or error-recovery behavior. Agent-R constructs revision trajectories by joining erroneous and successful branches discovered through Monte Carlo tree search, then trains agents to recover from mistakes \citep{yuan2025agent}. 
Agentic Critical Training instead rewards an agent for judging which of two candidate actions is better \citep{liu2026agentic},
and STeP trains on teacher-generated self-reflected trajectories containing both erroneous behavior and its correction \citep{chen2025training}. \method likewise trains on critiques and corrections, but its correction specifies what the agent should have done in place of the mistake, rather
than how to proceed afterward. Additionally, we explicitly isolate the supervision value of the critique itself, asking if it can better improves the agent’s task performance.

\section{Conclusion}
This work introduces \method, a critique-augmented SFT method for multiturn LLM
agents that fine-tunes the student on textual critiques and corrected actions generated by an expert over the student's own trajectories.
We find that critiques carry learning signal beyond the corrected action with \method-Critique outperforming \method-Correction and action-only distillation across all tested settings.
Training an agent to \emph{explain} its mistakes thus improves its ability to \emph{perform} the correct actions.
More broadly, \method exploits the fact that expert and student share a natural-language interface:
critiques are a cheap and rich form of feedback that can be incorporated directly into training.
We see them as one more axis along which LLM agents can be improved.

Several interesting directions follow from this work.
Our Gemma 4 31B results suggest that a model given privileged information can
improve from critiques of its own trajectories. Extending this self-critique setting to more models and domains is a natural next step. Because the critique
target is natural language, \method could also consume feedback from sources other than an expert LLM, most notably direct user feedback at deployment, which suggests online variants that train on corrections as they arrive.
Additionally, while our experiments with \method+BC show promise in combining critiques with expert demonstrations, identifying the precise conditions under which these mixtures succeed and exploring integrations with other supervision methods remain important areas for future study.
Finally, scaling to larger students and broader agentic domains, and
developing a theoretical account of why critique targets help, remain open.

\bibliography{iclr2027_conference}

@article{winata2026t1,
  title={T1-Bench: Benchmarking Multi-Scenario Agents in Real-World Domains},
  author={Winata, Genta Indra and Chakraborty, Amartya and Lin, Yuzhen and Rao, Swasthi P and Siingh, Shikhhar and Lu, Houhan and Bathaee, Nadia and Hatwar, Sriharsha and Dashore, Paresh and Jain, Anmol and others},
  journal={arXiv preprint arXiv:2606.11070},
  year={2026}
}

@inproceedings{
wang2025critique,
title={Critique Fine-Tuning: Learning to Critique is More Effective than Learning to Imitate},
author={Yubo Wang and Xiang Yue and Wenhu Chen},
booktitle={Second Conference on Language Modeling},
year={2025},
}

@inproceedings{
barres2026taubench,
title={\${\textbackslash}tau{\textasciicircum}2\$-Bench: Evaluating Conversational Agents in a Dual-Control Environment},
author={Victor Barres and Honghua Dong and Soham Ray and Xujie Si and Karthik R Narasimhan},
booktitle={Forty-third International Conference on Machine Learning},
year={2026},
}

@article{yao2022webshop,
  title={Webshop: Towards scalable real-world web interaction with grounded language agents},
  author={Yao, Shunyu and Chen, Howard and Yang, John and Narasimhan, Karthik},
  journal={Advances in Neural Information Processing Systems},
  volume={35},
  pages={20744--20757},
  year={2022}
}

@inproceedings{zeng2024agenttuning,
  title={Agenttuning: Enabling generalized agent abilities for llms},
  author={Zeng, Aohan and Liu, Mingdao and Lu, Rui and Wang, Bowen and Liu, Xiao and Dong, Yuxiao and Tang, Jie},
  booktitle={Findings of the Association for Computational Linguistics: ACL 2024},
  pages={3053--3077},
  year={2024}
}

@inproceedings{chen2024agent,
  title={Agent-flan: Designing data and methods of effective agent tuning for large language models},
  author={Chen, Zehui and Liu, Kuikun and Wang, Qiuchen and Zhang, Wenwei and Liu, Jiangning and Lin, Dahua and Chen, Kai and Zhao, Feng},
  booktitle={Findings of the Association for Computational Linguistics: ACL 2024},
  pages={9354--9366},
  year={2024}
}

@inproceedings{ross2011reduction,
  title={A reduction of imitation learning and structured prediction to no-regret online learning},
  author={Ross, St{\'e}phane and Gordon, Geoffrey and Bagnell, Drew},
  booktitle={Proceedings of the fourteenth international conference on artificial intelligence and statistics},
  pages={627--635},
  year={2011},
  organization={JMLR Workshop and Conference Proceedings}
}

@inproceedings{
kapusuzoglu2026critiqueguided,
title={Critique-Guided Distillation for Robust Reasoning via Refinement},
author={Berkcan Kapusuzoglu and Supriyo Chakraborty and Zain Sarwar and Chia-Hsuan Lee and Sambit Sahu},
booktitle={Forty-third International Conference on Machine Learning},
year={2026},
}

@inproceedings{
xu2026theagentcompany,
title={TheAgentCompany: Benchmarking {LLM} Agents on Consequential Real World Tasks},
author={Frank F. Xu and Yufan Song and Boxuan Li and Yuxuan Tang and Kritanjali Jain and Mengxue Bao and Zora Zhiruo Wang and Xuhui Zhou and Zhitong Guo and Murong Cao and Mingyang Yang and Hao Yang Lu and Amaad Martin and Zhe Su and Leander Melroy Maben and Raj Mehta and Wayne Chi and Lawrence Keunho Jang and Yiqing Xie and Shuyan Zhou and Graham Neubig},
booktitle={The Thirty-ninth Annual Conference on Neural Information Processing Systems Datasets and Benchmarks Track},
year={2025}
}

@inproceedings{mohammadi2025evaluation,
  title={Evaluation and benchmarking of llm agents: A survey},
  author={Mohammadi, Mahmoud and Li, Yipeng and Lo, Jane and Yip, Wendy},
  booktitle={Proceedings of the 31st ACM SIGKDD Conference on Knowledge Discovery and Data Mining V. 2},
  pages={6129--6139},
  year={2025}
}

@article{pan2025measuring,
  title={Measuring agents in production},
  author={Pan, Melissa Z and Arabzadeh, Negar and Cogo, Riccardo and Zhu, Yuxuan and Xiong, Alexander and Agrawal, Lakshya A and Mao, Huanzhi and Shen, Emma and Pallerla, Sid and Patel, Liana and others},
  journal={arXiv preprint arXiv:2512.04123},
  year={2025}
}

@inproceedings{chen2025atlas,
  title={Atlas: Agent tuning via learning critical steps},
  author={Chen, Zhixun and Li, Ming and Huang, Yuxuan and Du, Yali and Fang, Meng and Zhou, Tianyi},
  booktitle={Findings of the Association for Computational Linguistics: ACL 2025},
  pages={25334--25349},
  year={2025}
}

@article{li2026revisiting,
  title={Revisiting DAgger in the Era of LLM-Agents},
  author={Li, Changhao and Qiang, Rushi and Huang, Jiawei and Gao, Chenxiao and Zhang, Chao and He, Niao and Dai, Bo},
  journal={arXiv preprint arXiv:2605.12913},
  year={2026}
}

@inproceedings{agarwal2024policy,
  title={On-policy distillation of language models: Learning from self-generated mistakes},
  author={Agarwal, Rishabh and Vieillard, Nino and Zhou, Yongchao and Stanczyk, Piotr and Ramos Garea, Sabela and Geist, Matthieu and Bachem, Olivier},
  booktitle={International Conference on Learning Representations},
  volume={2024},
  pages={21246--21263},
  year={2024}
}

@article{yuan2025agent,
  title={Agent-r: Training language model agents to reflect via iterative self-training},
  author={Yuan, Siyu and Chen, Zehui and Xi, Zhiheng and Ye, Junjie and Du, Zhengyin and Chen, Jiecao},
  journal={arXiv preprint arXiv:2501.11425},
  year={2025}
}

@inproceedings{
yang2025lighthouse,
title={The Lighthouse of Language: Enhancing {LLM} Agents via Critique-Guided Improvement},
author={Ruihan Yang and Fanghua Ye and Jian Li and Siyu Yuan and Yikai Zhang and Zhaopeng Tu and Xiaolong Li and Deqing Yang},
booktitle={The Thirty-ninth Annual Conference on Neural Information Processing Systems},
year={2025}
}

@article{blakeman2025nvidia,
  title={Nvidia nemotron 3: Efficient and open intelligence},
  author={Blakeman, Aaron and Grattafiori, Aaron and Basant, Aarti and Gupta, Abhibha and Khattar, Abhinav and Renduchintala, Adi and Vavre, Aditya and Shukla, Akanksha and Bercovich, Akhiad and Ficek, Aleksander and others},
  journal={arXiv preprint arXiv:2512.20856},
  year={2025}
}

@article{team2026gemma,
  title={Gemma 4 technical report},
  author={Abd, Sherif El and Aggarwal, Vaibhav and Algayres, Robin and Andreev, Alek and Bachem, Olivier and Ballantyne, Ian and Brick, Cormac and C{\u{a}}rbune, Victor and Casbon, Michelle and others},
  journal={arXiv preprint arXiv:2607.02770},
  year={2026}
}

@inproceedings{
zhang2026advancing,
title={Advancing {LLM} Reasoning with Natural Language and Numerical Feedback},
author={Xiaoying Zhang and Yipeng Zhang and Hao Sun and Kaituo Feng and Chaochao Lu and Chao Yang and Helen M. Meng},
booktitle={Forty-third International Conference on Machine Learning},
year={2026},
}

@article{lin2026icrl,
  title={ICRL: Learning to Internalize Self-Critique with Reinforcement Learning},
  author={Lin, Jianbo and Yu, Xiaomin and Xin, Yi and Guo, Yifu and Jiang, Zhuosong and Yue, Zhongqi and Wang, Weishi and Zou, Heqing and Qin, Chengwei and Xiong, Hui},
  journal={arXiv preprint arXiv:2605.15224},
  year={2026}
}

@article{shinn2023reflexion,
  title={Reflexion: Language agents with verbal reinforcement learning},
  author={Shinn, Noah and Cassano, Federico and Gopinath, Ashwin and Narasimhan, Karthik and Yao, Shunyu},
  journal={Advances in neural information processing systems},
  volume={36},
  pages={8634--8652},
  year={2023}
}

@article{chen2025training,
  title={Training llm-based agents with synthetic self-reflected trajectories and partial masking},
  author={Chen, Yihan and Xu, Benfeng and Wang, Xiaorui and Zhang, Yongdong and Mao, Zhendong},
  journal={arXiv preprint arXiv:2505.20023},
  year={2025}
}

@article{liao2026multi,
  title={Multi-Turn On-Policy Distillation with Prefix Replay},
  author={Liao, Baohao and Dong, Hanze and Monz, Christof and Xu, Xinxing and Dong, Li and Wei, Furu},
  journal={arXiv preprint arXiv:2607.04763},
  year={2026}
}

@article{shao2024deepseekmath,
  title={Deepseekmath: Pushing the limits of mathematical reasoning in open language models},
  author={Shao, Zhihong and Wang, Peiyi and Zhu, Qihao and Xu, Runxin and Song, Junxiao and Bi, Xiao and Zhang, Haowei and Zhang, Mingchuan and Li, YK and Wu, Yang and others},
  journal={arXiv preprint arXiv:2402.03300},
  year={2024}
}

@inproceedings{choudhury2025better,
  title={Better than your teacher: LLM agents that learn from privileged AI feedback},
  author={Choudhury, Sanjiban and Sodhi, Paloma},
  booktitle={International Conference on Learning Representations},
  volume={2025},
  pages={10457--10493},
  year={2025}
}

@article{sun2026step,
  title={Step-Level On-Policy Distillation: Interpolating Between On-Policy Distillation and Supervised Fine-Tuning},
  author={Sun, Changhui and Liu, Lanbo and Lei, Hang and Ling, Tong and Xie, Jiahang and Zheng, Zhiyong and Wang, Yujia and Liu, Hao and Xiao, Feng and Liu, Lu and others},
  journal={arXiv preprint arXiv:2608.16333},
  year={2026}
}

@inproceedings{
ye2026few,
title={A Few Teacher Steps Go a Long Way: Cost-Efficient On-Policy Data Augmentation for Agent Post-Training},
author={Ye, Junze and Cheng, Jiayi and Lu, Miao and Mankowski, Michal and Blanchet, Jose and Bayati, Mohsen},
booktitle={ICML 2026 Workshop on RL from World Feedback},
year={2026},
}

@inproceedings{lyustudent,
  title={Student-Centered Distillation Narrows the Agentic Gap Between Small and Large LLMs},
  author={Lyu, Yuanjie and Wang, Chengyu and Huang, Jun and Xu, Tong},
  booktitle={Forty-third International Conference on Machine Learning},
  year={2026},
}

@article{an2023learning,
  title={Learning from mistakes makes llm better reasoner},
  author={An, Shengnan and Ma, Zexiong and Lin, Zeqi and Zheng, Nanning and Lou, Jian-Guang and Chen, Weizhu},
  journal={arXiv preprint arXiv:2310.20689},
  year={2023}
}

@inproceedings{ye2025physics,
  title={Physics of language models: Part 2.2, how to learn from mistakes on grade-school math problems},
  author={Ye, Tian and Xu, Zicheng and Li, Yuanzhi and Allen-Zhu, Zeyuan},
  booktitle={International Conference on Learning Representations},
  volume={2025},
  pages={78136--78147},
  year={2025}
}

@inproceedings{
zhang2026agent,
title={Agent Learning via Early Experience},
author={Kai Zhang and Xiangchao Chen and Bo Liu and Tianci Xue and Zeyi Liao and Zhihan Liu and Xiyao Wang and Yuting Ning and Zhaorun Chen and Xiaohan Fu and Jian Xie and Yuxuan Sun and Boyu Gou and Qi Qi and Zihang Meng and Jianwei Yang and Ning Zhang and Xian Li and Ashish Shah and Dat Huynh and Hengduo Li and Zi Yang and Xuefei Cao and Lawrence Keunho Jang and Shuyan Zhou and Jiacheng Zhu and Huan Sun and Jason E Weston and Yu Su and Yifan Wu},
booktitle={Forty-third International Conference on Machine Learning},
year={2026},
}

@article{saeidi2026cast,
  title={CAST: Critique-Aware Supervision for Training Reliable Long-Horizon Tool-Calling Agents},
  author={Saeidi, Amir and Zhang, Zehua and Singh, Rishitosh and Ahuja, Naman and Gupta, Vivek and Payani, Ali and Liu, Gaowen and Srinivasa, Jayanth and Baral, Chitta},
  journal={arXiv preprint arXiv:2608.30147},
  year={2026}
}

@article{liu2026agentic,
  title={Agentic critical training},
  author={Liu, Weize and Liu, Minghui and Ho, Sy-Tuyen and Chakraborty, Souradip and Wang, Xiyao and Huang, Furong},
  journal={arXiv preprint arXiv:2603.08706},
  year={2026}
}

@article{ouyang2022training,
  title={Training language models to follow instructions with human feedback},
  author={Ouyang, Long and Wu, Jeffrey and Jiang, Xu and Almeida, Diogo and Wainwright, Carroll and Mishkin, Pamela and Zhang, Chong and Agarwal, Sandhini and Slama, Katarina and Ray, Alex and others},
  journal={Advances in neural information processing systems},
  volume={35},
  pages={27730--27744},
  year={2022}
}

@article{guo2025deepseek,
  title={Deepseek-r1: Incentivizing reasoning capability in llms via reinforcement learning},
  author={Guo, Daya and Yang, Dejian and Zhang, Haowei and Song, Junxiao and Wang, Peiyi and Zhu, Qihao and Xu, Runxin and Zhang, Ruoyu and Ma, Shirong and Bi, Xiao and others},
  journal={arXiv preprint arXiv:2501.12948},
  year={2025}
}

@inproceedings{
wu2024agentkit,
title={AgentKit: Structured {LLM} Reasoning with Dynamic Graphs},
author={Yue Wu and Yewen Fan and So Yeon Min and Shrimai Prabhumoye and Stephen Marcus McAleer and Ruslan Salakhutdinov and Yonatan Bisk and Yuanzhi Li and Tom Mitchell},
booktitle={First Conference on Language Modeling},
year={2024},
url={https://openreview.net/forum?id=PKfAq8N4fK}
}

@article{oh2026neglected,
  title={Neglected Free Lunch from Post-training: Progress Advantage for LLM Agents},
  author={Oh, Changdae and Li, Wendi and Park, Seongheon and Yeh, Samuel and Mallick, Tanwi and Li, Sharon},
  journal={arXiv preprint arXiv:2606.26080},
  year={2026}
}

@inproceedings{sheng2025hybridflow,
  title={Hybridflow: A flexible and efficient rlhf framework},
  author={Sheng, Guangming and Zhang, Chi and Ye, Zilingfeng and Wu, Xibin and Zhang, Wang and Zhang, Ru and Peng, Yanghua and Lin, Haibin and Wu, Chuan},
  booktitle={Proceedings of the Twentieth European Conference on Computer Systems},
  pages={1279--1297},
  year={2025}
}
\bibliographystyle{iclr2027_conference}

\appendix

\section{Environments, Task Splits, and Evaluation Protocol}
\label{app:envs}

\paragraph{$\tau^2$-Bench Airline and Retail.}
$\tau^2$-Bench \citep{barres2026taubench} evaluates conversational agents that interact with a simulated
user while operating a domain database through tool calls and obeying a written domain policy.
We use the Airline and Retail domains, which we report jointly as $\tau^2$ AR.
The agent observes the domain policy and tool schema in its system prompt, and interacts with a user
simulator. Each task terminates when the agent ends the conversation or the turn budget is exhausted
($40$ agent turns). The benchmark uses a binary outcome grader, which checks the final
database state against the task's gold reference actions and evaluates any required user-facing
information with a natural-language judge. In our experiments, both the user simulator and the judge are Gemma 4 31B. We hold out $60$ tasks spanning both domains for evaluation and the remaining tasks form the collection pool. The data split follows the train and evaluation split specified in the official code release of $\tau^2$-bench (originally intended for people running RL training experiments). Each evaluation task is rolled out $16$ times and we report mean task success.

\paragraph{WebShop.}
WebShop \citep{yao2022webshop} places the agent in a simulated e-commerce site with a product
catalog. Given a natural-language product
request, the agent issues search queries, clicks results and option buttons, and finally purchases an
item.
Following the evaluation protocols of previous papers---cf.~the work of \cite{chen2024agent}, \cite{wu2024agentkit} and \cite{oh2026neglected}---we use the $1000$-product WebShop instance (\texttt{items\_shuffle\_1000.json}) fixing the first $100$ goals in the environment's fixed goal ordering as the evaluation set.  We further use the second $100$ goals to form the training pool. We run training and evaluation using the \texttt{text\_rich} observation mode.  Each evaluation goal is rolled out $16$ times. The agent is driven by a simple text-based ReAct harness where it receives the system prompt of Appendix~\ref{app:ws_prompts} once and then, at every turn, an \texttt{Observation:} message containing the rendered page with its clickable elements marked as \texttt{[button] \ldots [button\_]}. 
Each turn, the agent is required to end its turn with a single \texttt{Action: search[\ldots]} or \texttt{Action: click[\ldots]} line. An episode ends when the agent clicks \texttt{Buy Now} or after $30$ actions. We report the
fraction of episodes in which the purchased item satisfies all of the request's constraints (product type, attributes, options, and price limit).

\paragraph{Decoding Parameters.}
Trajectory collection for $\tau^2$ AR uses temperature $0.7$. WebShop uses temperature $1.0$ with
$\text{top-}p = 0.95$ for Nemotron and $\text{top-}k = 64$ for the Gemma students, matching their released
generation configurations.
Evaluation uses the same decoding parameters as collection.
The expert $C$ is Gemma 4 31B Instruct in all experiments and is queried with greedy decoding.

\section{Expert Annotation and Interjection Prompts}
\label{app:prompts}

We use different expert prompts for the two environments since their action spaces and reward structures are different. For $\tau^2$ AR, the expert judges tool calls and user-facing messages against a domain policy, whereas on WebShop it judges \texttt{search}/\texttt{click} actions against a single target product. Both prompts share the same design principles, following our \method method of giving the expert privileged information, having it judge one action step at a time and, for a flagged step, write a textual critique and the counterfactually corrected action. The student-side interjection $\rho$ (Appendix~\ref{app:interjection}) is identical in both environments.

\subsection{$\tau^2$ AR expert annotation prompt}
\label{app:tau2_prompts}

The expert is given, for each agent turn: the task description, the user's hidden scenario, the gold
reference actions together with the reward's evaluation criteria, an optional slice of the relevant
database state, the conversation so far, and the single step to be judged. It returns a criticality
flag, a verdict, and a free-text analysis. We use the analysis of a \textsc{misstep} verdict as the
critique $c_t$ and parse the text following ``\texttt{Instead: }'' as the corrected action
$\tilde a_t$. The system prompt body is given in Figure~\ref{fig:prompt-tau2-sys}. The short
per-domain note of Figure~\ref{fig:prompt-tau2-domain} is inserted immediately before item 1, and the
agent's full domain policy and tool schema are appended to it verbatim.

\begin{figure}[tp]
\begin{promptbox}[Expert annotation prompt --- system message]
\begin{promptverb}
You are a privileged COACH reviewing a customer-service AGENT. You review EVERY step the agent takes -- a
read-only lookup, a database write, or a message to the user -- AFTER it has already executed. You never
block, undo, or change anything; you only record a short verdict on the step taken.

You are given information the AGENT does NOT have: the GOLD reference actions, the reward's EVALUATION
CRITERIA, possibly a slice of the relevant DATABASE STATE, and the user's hidden SCENARIO -- plus the
agent's own full DOMAIN POLICY and TOOL SCHEMA (appended to these instructions). Use all of it to judge.

NOTE ON REASONING: the agent's message may include a PRIVATE chain-of-thought block that is NEVER shown to
the user. Judge ONLY the user-visible action (the text after that reasoning block, plus any tool call); do
not penalize the private reasoning, and do not treat it as a message sent to the user.

For the single step the agent just took, report TWO things:

1. CRITICALITY -- Using the gold reference actions and reward criteria you were given: a step is CRITICAL
   only if it directly performs, or is a necessary precondition for, a GOLD action / database write /
   assertion that the reward actually checks. Lookups, greetings, confirmations, rephrasings, and any step
   whose result the reward does not depend on are NOT critical, even if helpful. Answer YES or NO.

2. CORRECTNESS -- Judge the step against the gold reference actions, the reward criteria, and the agent's
   domain policy. Call it a genuine MISSTEP ONLY IF it diverges from what the gold reference and the policy
   require at THIS point in a way that affects the outcome: it acts on the wrong entity, skips a
   verification/precondition the policy mandates, performs an incorrect or irreversible write, or
   contradicts the policy or what the user actually said. If the step is consistent with the gold trajectory
   and the policy -- or is a harmless, recoverable, or cosmetic variation -- it is a PASS. Default HARD to
   PASS: reserve a misstep verdict for the FEW genuinely outcome-affecting errors; when in doubt, PASS.
   Judge CRITICALITY independently of correctness.
   For a misstep: state in ONE sentence what was wrong, then "Instead: " the concrete correct action. For a
   CRITICAL PASS: state in ONE sentence why it was the right step.

## Output format

Emit EXACTLY these three tags and nothing else. Do not think out loud.

<critical>YES or NO</critical>
<decision>PASS or MISSTEP</decision>
<analysis>
   For a MISSTEP: one sentence on what was wrong, then "Instead: " and the concrete counterfactual action as
   a NATIVE tool call (or the exact user instruction for a user-device tool). For a CRITICAL PASS: one
   sentence on why this was the correct step. For a non-critical PASS: may be a brief note or empty.

</analysis>

## Write the analysis in FIRST PERSON, with a LITERAL counterfactual

Write the one-sentence REASON as the agent reflecting on its own step, using "I"/"my" (e.g. "I skipped the
required verification before updating the reservation") -- never "the agent". After "Instead: " output ONLY
the LITERAL next action itself -- the exact text that would be emitted -- NOT a description of it.
\end{promptverb}
\end{promptbox}
\caption{\textbf{$\tau^2$-Bench expert annotation prompt: system message}. The per-domain note of Figure~\ref{fig:prompt-tau2-domain} is inserted immediately before item~1, and the agent's full domain policy and tool schema are appended to it verbatim.}
\label{fig:prompt-tau2-sys}
\end{figure}

\begin{figure}[htbp]
\begin{promptbox}[$\tau^2$-Bench domain notes --- inserted before item 1 (one per domain)]
\begin{promptverb}
[airline] This is the AIRLINE domain. The agent's database writes are reservation actions -- e.g.
book_reservation, cancel_reservation, update_reservation_flights, update_reservation_passengers,
update_reservation_baggage, send_certificate. Every tool is AGENT-callable; there are NO user-device tools,
so the agent should never tell the user to operate a tool themselves. Watch especially for acting on the
wrong reservation / flight / passenger, skipping the required confirmation before a paid change or
cancellation, or violating the cabin / baggage / change rules in the policy.

[retail] This is the RETAIL domain. The agent's database writes mutate orders -- cancel_pending_order,
modify_pending_order_items / _address / _payment, return_delivered_order_items,
exchange_delivered_order_items. Every tool is AGENT-callable; there are NO user-device tools, so the agent
should never tell the user to operate a tool themselves. Watch especially for acting on the wrong order /
item, exchanging to an unavailable or different-product variant, skipping the required user confirmation
before a write, or acting on an order in the wrong status (an order must be pending to cancel/modify,
delivered to return/exchange).
\end{promptverb}
\end{promptbox}
\caption{\textbf{$\tau^2$-Bench domain notes}, one per domain, inserted into the expert system message of Figure~\ref{fig:prompt-tau2-sys}.}
\label{fig:prompt-tau2-domain}
\end{figure}

In the prompt, we use a first-person directive asking the expert to generate the critique and corrected action so that $\tilde a_t$
is directly usable as a supervised target.
We adopt this as a practical design choice and do not ablate its specific impact on \method performance.
In the expert prompt, we additionally ask the expert to label the criticality of the step -- which we find is important to prevent it from overzealously labeling every step as incorrect. For our final \method runs, we only keep correct and incorrect steps that are critical.

\paragraph{Per-step context template.}
Each annotation query is completed with the user message of Figure~\ref{fig:prompt-tau2-user}. When privileged information is
withheld (Section~\ref{sec:critique_teacher}), only the \texttt{<conversation\_so\_far>} and \texttt{<step\_the\_agent\_just\_took>} blocks are kept. We remove every reference to the gold actions, reward criteria and scenario from the system prompt (Appendix~\ref{app:prompt_ablation}).

\begin{figure}[htbp]
\begin{promptbox}[Expert annotation prompt --- per-step user message]
\begin{promptverb}
<task_description>{...}</task_description>
<user_scenario>{...}</user_scenario>
<gold_reference_and_reward_criteria>
{...}
</gold_reference_and_reward_criteria>
<conversation_so_far>{...}</conversation_so_far>
<step_the_agent_just_took>{...}</step_the_agent_just_took>

Review the step above and emit EXACTLY <critical>YES|NO</critical>,
<decision>PASS|MISSTEP</decision>, and <analysis>...</analysis> as
specified -- nothing else.
\end{promptverb}
\end{promptbox}
\caption{\textbf{$\tau^2$-Bench expert annotation prompt: per-step user message.} Under the no-privilege condition only the \texttt{<conversation\_so\_far>} and \texttt{<step\_the\_agent\_just\_took>} blocks are kept.}
\label{fig:prompt-tau2-user}
\end{figure}

\subsection{WebShop expert annotation prompt}
\label{app:ws_prompts}
On WebShop the expert reviews each proposed \texttt{search[\ldots]} or \texttt{click[\ldots]} action given the shopping instruction, the privileged goal specification (namely the target product id, product type, required attributes and price limit), the clickable buttons on the agent's current page, and the episode so far. The expert system message is given in Figure~\ref{fig:prompt-ws-sys} and the per-step user message in Figure~\ref{fig:prompt-ws-user}. The student-side system prompt used by our WebShop harness is given in Figure~\ref{fig:prompt-ws-student}. The \method system addendum of Appendix~\ref{app:interjection} is appended to it during fine-tuning.

\begin{figure}[tp]
\begin{promptbox}[WebShop expert annotation prompt --- system message]
\begin{promptverb}
You are a COACH reviewing a WebShop shopping AGENT before its next action executes. The agent interacts with
a text store through two actions:
  - search[<query>]  - keyword search; works from ANY page (no need to click "Back to Search" first).
  - click[<button>]  - click a button CURRENTLY on the page: a product id, an option (color/size/...), "Buy
    Now", "Next >", "< Prev", "Back to Search", "Description", "Features", "Reviews".
Search returns the TOP-50 results (10 per page, up to 5 pages; "Next >" pages forward). The item that
satisfies the instruction is almost always somewhere in those 50 if the query names the specifics. To
succeed the agent must, IN ORDER: (1) issue a SPECIFIC query naming the product TYPE + key ATTRIBUTES + the
required color/size; (2) SCAN the results (page with "Next >") for the product that actually OFFERS the
required variant, not just a look-alike; (3) open that product; (4) select EVERY required OPTION; (5) click
"Buy Now" within the price limit. Searching and paging are FREE and expected.

You have privileged access the agent does NOT have:
- The target product and its required ATTRIBUTES, required OPTIONS (color/size/...), price limit, and type.
- The clickable BUTTONS on the agent's CURRENT page, and the queries/pages already seen in the history.

## BLOCK ONLY these, and always give a CONSTRUCTIVE FORWARD fix (never a bare "Back to Search"):
  S1. search[...] that is VAGUER than the instruction (drops the required color/size or a key attribute)
      when a more specific query is possible.
      -> Instead: Action: search[<product type + key attributes + required color + required size>]
  S2. search[...] that just REPEATS a query already tried this session (it returns the same results).
      -> If this page has "Next >": Action: click[Next >].  Else:
         Action: search[<reworded query that adds a distinguishing
         attribute or the exact color/size token>]
  O1. click[<product id>] opening a look-alike whose title/type clearly cannot match, when a
      clearly-matching listing is visible on THIS results page. -> Instead: Action: click[<the matching
      product id here>]
  P1. click[Buy Now] (or an option) on an opened product that does NOT offer the required color/size
      variant. Do NOT go back blindly:
      -> If this page has "Next >" (results remain unscanned): Action: click[Next >]
      -> Else: Action: search[<full instruction incl. the exact required color and size>]
  P2. click[Buy Now] on a product whose TYPE/ATTRIBUTES clearly don't match the instruction, or over the
      price limit. -> Instead:
      Action: click[Next >] (if results remain) or search[<more
      specific query>].
  B1. click[Buy Now] with a required OPTION still unselected AND that option's correct value is a button
      here.
      -> Instead: Action: click[<the exact correct option value>]
  B2. click[<option>] selecting the WRONG variant (a color/size contradicting a required option) when the
      correct value is a button here. -> Instead: Action: click[<the correct option value>]
  N1. An action that is NOT among THIS page's buttons at all (a guaranteed no-op).
      -> Instead: Action: <a valid forward button on this page>

## ALWAYS PASS (never BLOCK these):
- A FIRST, specific query; opening a plausibly-matching product to inspect it; selecting a CORRECT option;
  click[Buy Now] once every required option is selected on a matching product within price; and click[Next
  >] while unscanned pages remain. When unsure, PASS.

## The fix must be a REAL, PRESENT, FORWARD action:
- The fix target of a click MUST be copied verbatim from
  <current_page_buttons>. NEVER emit `Action: click[search]` and NEVER
  emit `Action: click[Back to Search]` as a fix -- to search again, emit `Action: search[...]` directly (it
  works from any page).
- Prefer, in order: select a required option here -> open the matching product here -> click[Next >] to scan
  more -> search[<more specific query>]. Only re-search when nothing on this page moves toward the match.

## No privileged framing - write ONLY what the agent could know from the screen
- NEVER use "gold", "target", "the correct answer", "the system knows"; never call a product id "the
  target".
- Justify from the OBSERVABLE instruction, e.g. "I searched without the requested color, so I only see
  look-alikes", "this product doesn't offer the requested size and I haven't scanned the remaining results",
  "I clicked Buy Now before selecting the requested color". Refer to requirements as the INSTRUCTION states.

## First person + literal action
- The critique is the AGENT talking about ITSELF ("I ...", never "the agent"), ONE sentence.
- After "Instead: " output ONLY the LITERAL next action: `Action: search[...]`, `Action: click[Next >]`,
  `Action: click[xx-large]`, `Action: click[B09KLQLLT2]`. NEVER intent/meta phrasing.

## Output format
Emit EXACTLY these tags and nothing else.
For PASS:
<decision>PASS</decision>
For BLOCK:
<decision>BLOCK</decision>
<analysis>One first-person sentence on what I did wrong. Instead:
Action: <the literal corrected action></analysis>
\end{promptverb}
\end{promptbox}
\caption{\textbf{WebShop expert annotation prompt: system message.} The expert sees the shopping instruction, the privileged goal specification, the buttons on the agent's current page, and the episode so far.}
\label{fig:prompt-ws-sys}
\end{figure}

\begin{figure}[htbp]
\begin{promptbox}[WebShop expert annotation prompt --- per-step user message]
\begin{promptverb}
<instruction>
{instruction}
</instruction>

<gold>
{gold}
</gold>

<current_page_buttons>
{buttons}
</current_page_buttons>

<history>
{history}
</history>

<proposed_action>
{action}
</proposed_action>

Review the proposed_action. Emit <decision>PASS</decision> to let it through, or <decision>BLOCK</decision>
plus a first-person <analysis> ending in "Instead: Action: ..." (the literal corrected action).
\end{promptverb}
\end{promptbox}
\caption{\textbf{WebShop expert annotation prompt: per-step user message.}}
\label{fig:prompt-ws-user}
\end{figure}

\begin{figure}[tp]
\begin{interjbox}[WebShop student system prompt (ReAct harness)]
\begin{promptverb}
You are an autonomous shopping agent operating a text-based web store (WebShop). Your job: buy the ONE
product that best matches the given Instruction.

Each turn, first reason about the current observation, then end your reply with EXACTLY ONE action line:
  Action: search[<query>]   - search the store; works from ANY page
  (you do NOT need to click "Back to Search" first)
  Action: click[<button>]   - click a button shown in the observation
  between [button] and [button_] (product ids like B08KBVJ4XN, options
  like "white"/"x-large", "Buy Now", "Description", "Next >", "Back to Search")

Guidelines:
- ONE ACTION PER TURN: after your reasoning, output a SINGLE "Action: ..." line and then STOP. Do NOT emit
  multiple "Action:" lines and do NOT write your own "Observation:" lines - only your FIRST action is
  executed and everything after it is ignored (so a plan written as several actions loses all but the
  first).
- Copy the button text EXACTLY as shown between [button] and [button_].
- Normal flow: search[...] -> click a product id -> click EACH required option (color/size/scent/...) ->
  click[Buy Now].
- Satisfy the WHOLE Instruction: every attribute, every option, and the price limit (price must be <= the
  stated dollars). Prefer the cheapest item that satisfies everything.
- If the system says your action had no effect, you are on the wrong page or used a wrong label: click the
  product id again to return to the item page, or "Back to Search" to search again.
- Make progress each turn; click Buy Now once your chosen item is configured to match.

Example:
Observation:
Instruction:
i would like a 3 ounce bottle of bright citrus deodorant for sensitive skin, and price lower than 50.00
  dollars
[button] Search [button_]
Action: search[3 ounce bright citrus deodorant sensitive skin]
Observation:
[button] Back to Search [button_]
Page 1 (Total results: 50)
[button] B078GWRC1J [button_]
Bright Citrus Deodorant by Earth Mama, 3-Ounce
$10.99
[button] B08KBVJ4XN [button_]
Barrel and Oak Aluminum-Free Deodorant, 2.7 oz
$15.95
Action: click[B078GWRC1J]
Observation:
[button] Back to Search [button_]
scent [button] bright citrus [button_] [button] ginger fresh [button_]
size [button] travel set (4-pack) [button_] [button] 3 ounce (pack of 1) [button_]
Price: $10.99
[button] Buy Now [button_]
Action: click[bright citrus]
Observation: You have clicked bright citrus.
Action: click[3 ounce (pack of 1)]
Observation: You have clicked 3 ounce (pack of 1).
Action: click[Buy Now]
\end{promptverb}
\end{interjbox}
\caption{\textbf{WebShop student system prompt} used by our ReAct harness. The \method{} system addendum of Figure~\ref{fig:interj-caft} is appended to it during fine-tuning.}
\label{fig:prompt-ws-student}
\end{figure}
\subsection{Prompt-ablation variants}
\label{app:prompt_ablation}

The ablation of Section~\ref{sec:critique_teacher} varies two axes independently on $\tau^2$ AR.
\emph{Privileged information} controls whether the expert receives the gold reference actions,
reward criteria and hidden user scenario. When they are withheld, the opening paragraph of the system prompt is replaced by the one in Figure~\ref{fig:prompt-noprivilege} and the correctness and criticality clauses drop their references to the gold reference and reward. \emph{Critique guidance} controls the elaborateness of the domain note and the criticality and
correctness clauses. The detailed guidance version uses the text of Figure~\ref{fig:prompt-tau2-sys}, while the
``minimal'' condition drops the domain note and replaces both clauses with the bare questions of
Figure~\ref{fig:prompt-minimal}, keeping the output format and the first-person directive
identical.

\begin{figure}[htbp]
\begin{minpromptbox}[No-privilege opening paragraph --- replaces the first two paragraphs of the system prompt]
\begin{promptverb}
You are an expert REVIEWER of a customer-service AGENT. You review EVERY step the agent takes -- a read-only
lookup, a database write, or a message to the user -- AFTER it has already executed. You never block, undo,
or change anything; you only record a short verdict on the single step taken.

You are NOT given any privileged information: you do NOT see a gold reference, the reward's evaluation
criteria, the user's hidden scenario, or the database state. Judge using ONLY the conversation so far and
the agent's own DOMAIN POLICY and TOOL SCHEMA (appended to these instructions) -- exactly the information
the agent itself has when it acts.
\end{promptverb}
\end{minpromptbox}
\caption{\textbf{No-privilege opening paragraph}, which replaces the first two paragraphs of the expert system message of Figure~\ref{fig:prompt-tau2-sys}.}
\label{fig:prompt-noprivilege}
\end{figure}
\begin{figure}[htbp]
\begin{minpromptbox}[Minimal critique guidance --- replaces criticality and correctness clauses 1 and 2]
\begin{promptverb}
1. CRITICALITY -- Does this step matter for whether the task ultimately succeeds? Answer YES or NO.

2. CORRECTNESS -- Was the step a mistake? If it was a genuine error, say MISSTEP; otherwise say PASS. For a
   misstep: state in ONE sentence what was wrong, then "Instead: " the concrete correct action. For a
   CRITICAL PASS: state in ONE sentence why it was the right step.
\end{promptverb}
\end{minpromptbox}
\caption{\textbf{Minimal critique guidance}, which replaces criticality and correctness clauses 1 and 2 of Figure~\ref{fig:prompt-tau2-sys}; the output format and the first-person directive are unchanged.}
\label{fig:prompt-minimal}
\end{figure}
All four conditions re-annotate the \emph{same} student trajectories with the same expert model, so
the comparison isolates the annotation prompt.

\subsection{Interjection $\rho$ and ablation variants}
\label{app:interjection}

The interjection $\rho$ is a fixed user-role message inserted after an expert-flagged action. The
student's system prompt additionally carries a short addendum disclosing that a coach is reviewing its
actions, so that the interjection is in-distribution during SFT training.
We crucially note that both these interjections are only used at \emph{training} time. During evaluation, we never use the expert critique to interject in any of its generations.
\method and its two
ablations differ only in these two strings (Figures~\ref{fig:interj-caft}--\ref{fig:interj-critique}) and in the corresponding target $y_t$.

\begin{figure}[htbp]
\begin{interjbox}[\method{} (full) --- system addendum and interjection $\rho$]
\begin{promptverb}
[system addendum]
A privileged coach silently reviews each action you take. If it judges a step you just took to be a mistake,
your NEXT turn will be prefixed with [COACH-FLAG]. When you see [COACH-FLAG], reconsider: work out what was
wrong with the action you just took and what the correct step is.

[interjection]
[COACH-FLAG] The coach flagged the action you just took as a mistake. Explain what specifically was wrong
with that action, then take the correct action instead.
\end{promptverb}
\end{interjbox}
\caption{\textbf{\method{} (full)}: student system addendum and interjection $\rho$.}
\label{fig:interj-caft}
\end{figure}

\begin{figure}[htbp]
\begin{interjbox}[\method-Correction --- system addendum and interjection $\rho$]
\begin{promptverb}
[system addendum]
A privileged coach silently reviews each action you take. If it judges a step you just took to be a mistake,
your NEXT turn will be prefixed with [COACH-FLAG]. When you see [COACH-FLAG], simply take the correct action
instead.

[interjection]
[COACH-FLAG] The coach flagged the action you just took as a mistake. Take the correct action instead.
\end{promptverb}
\end{interjbox}
\caption{\textbf{\method-Correction}: student system addendum and interjection $\rho$.}
\label{fig:interj-correction}
\end{figure}

\begin{figure}[htbp]
\begin{interjbox}[\method-Critique --- system addendum and interjection $\rho$]
\begin{promptverb}
[system addendum]
A privileged coach silently reviews each action you take. If it judges a step you just took to be a mistake,
your NEXT turn will be prefixed with [COACH-FLAG]. When you see [COACH-FLAG], state in your response what 
specifically was wrong with the action you just took.

[interjection]
[COACH-FLAG] The coach flagged the action you just took as a mistake. Explain what
specifically was wrong with that action.
\end{promptverb}
\end{interjbox}
\caption{\textbf{\method-Critique}: student system addendum and interjection $\rho$.}
\label{fig:interj-critique}
\end{figure}

\section{Data Collection and Supervised Fine-Tuning Details}
\label{app:sft}

\subsection{Trajectory collection and annotation}

For each student and environment we roll out the base policy once over the training task set with
$16$ rollouts per task, recording every agent turn together with its expert annotation.

All models are fully fine-tuned in bf16 with FSDP on a single $8$-GPU node
(NVIDIA RTX PRO 6000, $96$\,GB), using AdamW, gradient clipping at $1.0$, and right truncation.
For all \method, ablation, and action distillation experiments, we use a batch size of 64 and a learning rate of $3 \times 10^{-6}$. We train for 2 epochs on $\tau^2$ AR and 1 epoch on WebShop. For expert distillation and \method+BC mixtures on $\tau^2$ AR, we increase the learning rate to $1 \times 10^{-5}$ for 2 epochs. Iteration-2 \method retains the same hyperparameters of the first iteration.

The \method+BC mixtures of Section~\ref{sec:hybrid_distillation} are expert model compute-matched. Each mixture
contains a fixed total of $1664$ trajectory-equivalents, split between expert trajectories and
\method examples according to the stated ratio.

\section{Reinforcement Learning Details}
\label{app:rl}

We use GRPO as implemented in \texttt{verl} \citep{sheng2025hybridflow}. The policy self-serves through vLLM on the training GPUs, while the user simulator and
outcome judge run on an external Gemma 4 31B endpoint.
Each RL run occupies one $8$-GPU node. The reward is the binary $\tau^2$ outcome grader applied at
the end of each episode.
Table~\ref{tab:rl_hparams} lists the hyperparameter settings. Only the initialization differs across the three runs
of Section~\ref{sec:rl_warmstart}.

\begin{table}[h]
\centering
\small
\caption{GRPO hyperparameters for the $\tau^2$ AR warm-start experiments using Nemotron 3 Nano 4B.}
\label{tab:rl_hparams}
\begin{tabular}{lll}
\toprule
 Parameter & Value & \\
\midrule
Advantage estimator & GRPO & \\
Rollouts per prompt ($n$) & 16 & \\
Prompts per step & 32 & \\
Mini-batch size & 16 & \\
Micro-batch size per GPU & 1 & \\
Learning rate & $5\times10^{-7}$ & \\
LR schedule & constant, no warmup & \\
PPO clip range & $0.2$ & \\
KL loss & low-variance, coeff. $10^{-3}$ & \\
KL in reward & no & \\
$\gamma$, $\lambda_{\text{GAE}}$ & $1.0$, $1.0$ & \\
Entropy coefficient & $0$ & \\
Loss aggregation & token-mean & \\
Gradient clipping & $1.0$ & \\
Rollout temperature & $0.6$ & \\
Max prompt / response tokens & $12288$ / $16384$ & \\
Max agent / user turns & $100$ / $100$ & \\
Total steps & $300$ & \\
\bottomrule
\multicolumn{3}{l}{\footnotesize}
\end{tabular}
\end{table}

\section{Additional Results}
\label{app:extra}

\subsection{How much should correct turns be supervised?}
\label{app:passweight}

Section~\ref{methods} retains expert-approved turns as ordinary behavior-cloning targets, weighted by a scalar $\lambda$ relative to the corrective turns (we use $\lambda=0.1$ throughout, originally chosen from a sweep on a small training subset). In this section we evaluate a sweep of $\lambda \in \{0.01, 0.1, 0.5\}$ on the $\tau^2$ AR validation data for Gemma 4 E2B and Nemotron 3 Nano 4B.

Table~\ref{tab:passweight} shows that $\lambda=0.1$ is best for both students. Upweighting correct turns to $\lambda=0.5$ costs $7.2$\% and $5.6$\% for Gemma 4 E2B and Nemotron 3 Nano 4B, respectively, and nearly removing them ($\lambda=0.01$) costs $0.8$\% and $7.3$\%. We believe that at
$\lambda=0.5$ the objective is dominated by imitating the student's own executed actions, rather than on the critiques of incorect steps. However, removing the correct turns altogether is
even more damaging. Masking them out entirely costs $7.1$\% for Nemotron 3 Nano 4B
($53.2\%$ versus $60.3\%$ for a matched $\lambda=0.1$ run of the same configuration).
Without any correct-turn loss the model
sees only mistake-conditioned targets and degrades on the ordinary, correct turns that make up the
large majority of a trajectory.

\begin{table}[h]
\centering
\small
\caption{
$\tau^2$ AR task success (\%) as a function of the correct-turn loss weight $\lambda$
(\method uses $\lambda=0.1$ everywhere else in the paper).
All arms share the same student trajectories and expert annotations and differ only in $\lambda$.
}
\label{tab:passweight}
\begin{tabular}{lrrr}
\toprule
Student Model
    & $\lambda=0.01$
    & $\lambda=0.1$
    & $\lambda=0.5$ \\
\midrule

 Gemma 4 E2B & 45.8 & \textbf{46.6} & 39.4 \\
 Nemotron 3 Nano 4B & 53.0 & \textbf{60.3} & 54.7 \\

\bottomrule
\end{tabular}
\end{table}

\subsection{Mixing ratio for \method+BC}
\label{app:mixing}

Section~\ref{sec:hybrid_distillation} reports a $50/50$ mixture of expert trajectories and \method
supervision. Table~\ref{tab:mixing} sweeps the ratio at matched total training compute. An equal mix
is best for both Gemma students, while Nemotron---the only student for which \method alone already
beats expert distillation---prefers the mixture that contains the \emph{most} expert critique data. 
This is consistent with expert trajectories being complementary
only when behavior cloning on them is itself effective for the student 

\begin{table}[h]
\centering
\small
\caption{$\tau^2$ AR task success (\%) as a function of the expert-trajectory fraction in
\method+BC.
The $0\%$ column is pure \method and the reference for the sweep.}
\label{tab:mixing}
\begin{tabular}{lrrrr}
\toprule
Student & $0\%$ (\method) & $25\%$ BC & $50\%$ BC & $75\%$ BC \\
\midrule
Gemma 4 E2B & 46.6 & 47.2 & \textbf{53.7} & 51.5 \\
Gemma 4 E4B & 52.1 & 56.3 & \textbf{60.1} & 52.8 \\
Nemotron 3 Nano 4B & \textbf{60.3} & 56.7 & 56.0 & 56.5 \\
\bottomrule
\end{tabular}
\end{table}

\subsection{Do \method-trained agents receive less intervention from the expert model?}
\label{app:expert_interventions}

\begin{table}[h]
\centering
\small
\caption{\textbf{Expert Interventions.} Task success (\%) when the expert corrects flagged actions at inference time, alongside the average number of expert corrections per episode (Corr./ep.).}
\label{tab:interventions}
\begin{tabular}{llrrrrrr}
\toprule
& & \multicolumn{2}{c}{Base} & \multicolumn{2}{c}{\method} & \multicolumn{2}{c}{\method+BC} \\
\cmidrule(lr){3-4} \cmidrule(lr){5-6} \cmidrule(lr){7-8}
Environment & Student & Succ. & Corr./ep. & Succ. & Corr./ep. & Succ. & Corr./ep. \\
\midrule
\multirow{3}{*}{$\tau^2$ AR} 
& Gemma 4 E2B & 66.7 & 4.19 & 70.3 & 3.20 & 70.4 & 1.19 \\
& Gemma 4 E4B & 71.4 & 2.51 & 76.1 & 2.08 & 71.4 & 1.11 \\
& Nemotron 3 Nano 4B & 77.9 & 2.55 & 81.1 & 2.15 & 69.9 & 1.32 \\
\midrule
\multirow{3}{*}{WebShop} 
& Gemma 4 E2B & 80.0 & 4.77 & 72.0 & 3.73 & --- & --- \\
& Gemma 4 E4B & 80.0 & 5.25 & 68.0 & 2.82 & --- & --- \\
& Nemotron 3 Nano 4B & 81.0 & 5.99 & 75.0 & 2.66 & --- & --- \\
\bottomrule
\end{tabular}
\end{table}

To provide further evidence that \method effectively teaches the agent to avoid previously identified mistakes, we run an additional experiment where we 
take the \method-trained model checkpoints and measure the frequency of expert interventions that would take place if the expert was available at inference time.
We hypothesize that if the student policy successfully internalizes the textual critiques provided during training, it should naturally learn to avoid those same errors that the expert model typically flags and thus trigger fewer corrections from the experts than the base untrained model.

We test this hypothesis by conducting a coach-in-the-loop rollout on the validation split. During this evaluation, the expert model reviews every action taken by the student and replaces it with a corrected action whenever a mistake is flagged. We track both the resulting corrected task success rate and the total number of interventions required. 

As shown in Table~\ref{tab:interventions}, \method checkpoints require substantially fewer interventions compared to their corresponding base policies across both environments. On $\tau^2$ AR, this reduction in necessary corrections corresponds to a higher corrected success rate. The \method+BC checkpoints require the fewest interventions overall, though their corrected success rates indicate that the few remaining errors they make are more difficult to recover from. On WebShop, the \method policies similarly require far fewer corrections than the base models. Although similarly the actual success rate with intervention drops -- suggesting that while \method adapted to prevent the errors the expert critique does catch, the errors not caught by the expert become more fatal.
Ultimately, this drop in required expert interventions from the base model to the \method trained version supports the conclusion that \method agents successfully internalize the critique feedback.

\end{document}